# libtissue - implementing innate immunity

Jamie Twycross, Uwe Aickelin

*Abstract*— In a previous paper the authors argued the case for incorporating ideas from innate immunity into artificial immune systems (AISs) and presented an outline for a conceptual framework for such systems. A number of key general properties observed in the biological innate and adaptive immune systems were highlighted, and how such properties might be instantiated in artificial systems was discussed in detail. The next logical step is to take these ideas and build a software system with which AISs with these properties can be implemented and experimentally evaluated. This paper reports on the results of that step - the `libtissue` system.

## I. INTRODUCTION

`libtissue` is a software system for implementing and evaluating AIS algorithms on real-world monitoring and control problems. AIS algorithms are implemented as multi-agent systems of cells, antigen and signals interacting within tissue compartments. Input data is provided by sensors which monitor a system under surveillance, and cells are actively able to affect the monitored system through response mechanisms. `libtissue` provides a *general* implementational framework within which many different AIS algorithms can be instantiated, rather thanc [1]. `libtissue` is being used at the University of Nottingham to explore the application of a range of novel immune-inspired algorithms to problems in intrusion detection.

A brief review of the biological and conceptual views that underpin the design of `libtissue` is given in Section II, more detailed background information can be found in a previous paper [2]. This is then followed by a detailed description of the `libtissue` implementation in Section III. `libtissue` has grown into a fairly complex software system and its use is better understood in the context of examples. Thus, Section IV shows how `libtissue` can be applied to a real-world problem in computer security, and Section V describes the implementation of a simple example algorithm using `libtissue`. An analysis and evaluation of this algorithm are then presented in Section VI. The paper concludes with a brief summary and discussion of future work in Section VII.

## II. APPLYING INNATE IMMUNITY

In a previous paper [2] the authors describe several biological processes in detail and then discuss these biological processes at a conceptual level. This biological and conceptual view of the immune system forms the foundation upon which the `libtissue` implementation is built, and a brief summary is given here. The reader is referred to [2] and [3] for further discussions and explanations of the biological terminology.

Jamie Twycross, jpt@cs.nott.ac.uk (corresponding author), and Uwe Aickelin, uxa@cs.nott.ac.uk, are at the University of Nottingham, U.K.

The biological immune system is a complex system of cells of different types interacting with each other and the tissue in which they reside. The key elements of the system are cells, signals and antigen, combined with the environment, tissue. Cells have access to their environment through antigen and signals. Essentially, signals provide cells with information on the *behaviour* of entities in their environment, while antigen provides cells with information on the *structure* of these entities. In the biological system structure reflected at an antigenic level and behaviour at a signal level are tightly coupled. If the behaviour of a cell changes then so does its antigen profile and vice versa. Part of the motivation for the research presented here comes from a desire to better understand how information from these two levels determines the dynamics of the immune system.

As well as providing information on behaviour, signals also provide a control mechanism for immune system cells. The behaviour of a single cell is determined by complex signalling networks which are actively maintained between cells. A cell's behaviour can be seen in terms of the functions a cell performs. Of particular interest are the functions of antigen processing, signal processing, cellular binding, antigen matching and antigen response. Simple antigen processing consists of two steps: antigen ingestion and antigen presentation. During ingestion, antigen is transferred from the extracellular space to the interior of the cell. During presentation, internalised antigen is displayed on the surface of the cell. Additional manipulation of the antigen whilst inside the cell is also possible. A specialised class of cells called APCs performs antigen processing in the body. Signal processing refers to the ability of a cell to have its behaviour influenced through the level of a signal, such as a cytokine or hormone in the extracellular space. Control of DCs by PAMPs and Danger Signals, or of T helper cells by DCs provide good examples of this.

While signals allow cells to influence each other without coming into contact, many immune system processes involve interactions between cells which require contact. Cells bind with each other through the action of adhesion molecules and receptors on their surfaces. Antigen matching, the ability of certain classes of receptors, for example TcRs, to only be activated by specific patterns of antigen is one example of this. Antigen matching within a particular context leads to cells mounting a response, such as the initiation of the complement cascade. This response has an impact on the environment, causing other cells to change their behaviour, and so their structure, and closes the loop between cell and environment.

## III. SYSTEM IMPLEMENTATION

The aim of the research presented here is to build a software system which allows researchers to implement and analyse novel AIS algorithms and apply them to real-world problems. This clearly translates into three separate areas of functionality: algorithm implementation, algorithm analysis and algorithm application. This section begins by describing how the overall architecture of `libtissue` delivers these functionalities. It then goes on to present in as much technical detail as space permits how these functionalities have actually been implemented.

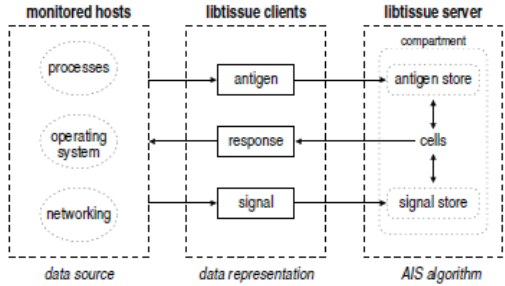

Fig. 1. The architecture of `libtissue`. `libtissue` clients monitor a host and provide input data to a `libtissue` server and AIS algorithm. Clients also allow algorithms to change the state of the monitored host.

`libtissue` has a client/server architecture pictured in Figure 1. An AIS algorithm is implemented as part of a `libtissue` server, and `libtissue` clients provide input data to the algorithm and response mechanisms which change the state of the monitored system. This client/server architecture separates data collection by the `libtissue` clients from data processing by the `libtissue` servers and allows for relatively easy extensibility and testing of algorithms on new data sources. `libtissue` is coded in C as a Linux shared library with client and server APIs, allowing new antigen and signal sources to be easily added to `libtissue` servers from a programmatic perspective. Because `libtissue` is implemented as a library, algorithms can be compiled and run on other machines with no modification. Client/server communication is socket-based, allowing clients and servers to potentially run on separate machines, for example a signal or antigen client may in fact be a remote network monitor.

AIS algorithms are implemented within a `libtissue` server as multiagent systems of cells. Cells exist within an environment, called a tissue compartment, along with other cells, antigen and signals. The problem to which the algorithm is being applied is represented by `libtissue` as antigen and signals. Cells express various repertoires of receptors and producers which allow them to interact with antigen and control other cells through signalling networks. `libtissue` allows data on implemented algorithms to be collected and logged, allowing for experimental analysis of the system.

### A. libtissue clients

`libtissue` clients are of three types: antigen, signal and response. Antigen clients collect and transform data into antigen which are forwarded to a `libtissue` server. Currently, a `systrace` antigen client has been implemented which collects process system calls (syscalls) using `systrace` [4]. Syscalls are a low-level mechanism by which applications request system services such as peripheral I/O or memory allocation from an operating system. Signal clients monitor system behaviour and provide an AIS running on the tissue server with input signals. A process monitor signal client, which monitors a process and its children and records statistics such as CPU and memory usage, and a network signal client, which monitors network interface statistics such as bytes per second, have been implemented. Two response clients have been implemented, one which simply logs an alert, and another which allows an active response through the modification of a `systrace` syscall policy. All of these clients are designed to be used in realtime experiments and for data collection for offline experiments with `tcreplay`.

The implementation is designed to allow varied AIS algorithms to be evaluated on real-world, realtime systems and problems. When testing IDSs it is common to use preexisting datasets such as the Lincoln Labs dataset [5]. However, the project `libtissue` has been built for is focused on combining measurements from a number of different concurrent data sources. Preexisting datasets which contain all the necessary sources are not available. Therefore, to facilitate experimentation, a `libtissue` replay client, called `tcreplay`, was also implemented. This client reads in log files gathered from previous realtime runs of antigen and signal clients, and also has the facility to read logfiles generated by `strace` [6]. It then sends these logs to a `libtissue` server. Variable replay rates are available, allowing data collected from a realtime session to be used to perform many experiments quickly. Having such a replay facility is important in terms of reproducibility of experiments. In this paper, all experimental runs are scripts which take data and parameter files as input and run a tissue server and `tcreplay` client.

### B. libtissue servers

A `libtissue` server is in fact several threaded processes running asynchronously. An initialisation routine is first called which creates a tissue compartment based on user-supplied parameters. During initialisation a thread is also started to handle connections between the server and `libtissue` clients, and this thread itself starts a separate thread for each connected client. After initialisation, cells, the characteristics of which are specified by the user, are created and initialised, and the tissue compartment populated with these cells. Cells in the tissue compartment then cycle and input data is provided by connected `libtissue` clients.

*1) Tissue compartments:* The `libtissue` server provides a multiagent simulation engine in which AIS algorithms can be implemented. At the centre of this simulation is the concept of a tissue compartment. A tissue compartment is the environment in which cells, signals and antigen interact. As well as housing cells, the maximum number of which is determined by the max_cells parameter, each tissue compartment has a fixed-size antigen store, set by the max_antigen parameter, where antigen provided by `libtissue` clients is placed. The tissue compartment also stores a fixed-number of signals, set by the max_cytokines parameter, the levels of which are set either by signal tissue clients or cells.

Input data can undergo some preprocessing before entering a tissue compartment. As well as representing the target-domain problem as antigen and signals, one of the roles of `libtissue` is to frame it in a "*more biological*" way in the following sense. The biological systems which biologically-inspired algorithms are based upon are specific for a particular environment with particular characteristics. For the immune system these characteristics include rate of antigen, uniqueness of antigen and antigen turnover. `libtissue` implements these functions by allowing the preprocessing of data from `libtissue` clients before it enters the tissue, controlled by a number of user-defined parameters. The antigen_multiplier parameter determines the number of copies of an incoming antigen placed into the tissue antigen storage. It was found necessary to have such a parameter since, as will be seen, datum in real-world problems often occur at a low frequency. Biologically, it is the case that a certain level of antigen is necessary to simulate the system, this is, a single unique antigen will not perturb the immune system much. Seen from the level of the pathogen, which is made up of repeated protein structures and reproduces itself multiple times, this is also clear. The multiplicity of antigen seems to be an important property of the biological immune system. In essence, the antigen_multiplier parameter allows `libtissue` to emulate this property for problems which have differing degrees of multiplicity in their input data, and its value is therefore problem dependent.

Another important concept related to antigen multiplicity is that of antigen persistence. In the biological system individual antigen do not persist indefinitely, but instead there is a turnover of antigen. This is provided for by `libtissue` on one level through the limitation of the size of a tissue compartment's antigen store by the max_antigen parameter, and can be seen by tracing the transit of antigen that are received by a `libtissue` server. After preprocessing by the `libtissue` server as detailed above, new antigen, multiplied by the antigen_multiplier parameter, will simply overwrite existing antigen. Antigen is then transferred from the tissue to the internal antigen store of cells with antigen receptors. From the internal store antigen is transferred to antigen producers on these cells, where it persists for a user-defined time period before being removed. Users can also remove antigen from a cell's store in the cell cycle callback. These factors combine to create a turnover of antigen in the tissue, with antigen entering and eventually being removed.

Even when composed of relatively small numbers of simple actors, the behaviour of multiagent systems is often difficult to understand. While formal analysis is possible, an experimental approach is more often adopted. `libtissue` implements probes which periodically sample and log data from a tissue compartment. Sampling is necessary, since even with simple algorithms such as the one described in Section V below it is infeasible in terms of storage space and performance to log all of the data produced. Additionally, since any experiment will require only certain data, the details of what is logged are left to the user, who provides a probe callback function. The rate at which this callback is run, and so the rate at which data is sampled, is defined by the probe_rate parameter. Probes allow data to be efficiently gathered and ease the experimental evaluation of algorithms.

*2) libtissue cells:* `libtissue` cells, like tissue compartments, have antigen and signal stores, the sizes of which are set by the num_antigen and num_cytokines parameters. They also have a number of different receptors and producers which allow them to interact with others cells, antigen and signals in the tissue compartment. Currently, four types of receptors have been implemented: antigen, cytokine, cell and VR receptors. Antigen receptors allow cells to transfer antigen from the tissue compartment to their own internal antigen store. Cytokine receptors allow cells to read signal levels in the compartment. Cell receptors allow cells to bind to other cells. Binding is necessary for VR receptors to be activated, which match antigen presented on another cell. Antigen from a cell's internal store are presented on antigen producers, one of the three types of producers currently implemented. The other two types, response and cytokine producers, allow cells to communicate with response clients, and to change signal levels in the tissue compartment and hence control the behaviour of cells with cytokine receptors respectively.

While `libtissue` provides the basic building blocks for modelling biological cells in terms of receptors and producers, the details of their actual configuration on cells and how cells behave in response to them is specified by the user. `libtissue` implements a simple scheduler which is periodically called at a rate defined by the cell_update_rate parameter. When called, the scheduler, taking the cells in a random order, first sets the values of the receptors for all of the cells. A user-defined cell cycle callback function is then executed for each cell. This function is essentially the controller for the cell, and determines how the actions of its receptors and producers are related. Once all of the callbacks have been run, the scheduler updates the tissue compartment according to cells' antigen producers. This design, since the cell cycle callback is in fact an arbitrary C function, means that cells can have complex behaviours. The specific action and parameters of the various producers and receptors is now described in more detail.

Antigen receptors allow the transfer of antigen from the tissue compartment's antigen store to the internal store of a cell. Transferred antigen is removed from the tissue compartment. For each antigen receptor a cell has, a random location in the tissue antigen store is chosen. If the location contains no antigen then none is transferred for that receptor.

A random location is picked in the cell's antigen store into which to transfer the tissue antigen. If this location contains a previously transfered antigen then it is overwritten by the incoming antigen. Clearly, both the parameter settings for the size of the tissue and cell antigen compartments, max_antigen and num_antigen respectively, as well as the rate of incoming antigen to the tissue compartment, will affect the overall rate at which antigen is transferred from the tissue compartment to the internal antigen store of cells.

Cytokine receptors allow cells to read the values of the signals stored in the tissue compartment, which are set by `libtissue` signal clients or cells themselves through cytokine producers. As well as providing a control mechanism for cells, cytokine receptors are designed to give cells sensitivity to external signals. Cytokine receptors are initialised as receptive to a specific tissue cytokine and at each iteration the value of this cytokine is copied to the receptor. This value is available for use during the user-specified cell cycle callback, and can, for example, affect the value of an internal cytokine or be used to determine the range of receptors a cell expresses.

Cell receptors model the concept of cellular binding in `libtissue` and enable cells to restrict some receptor/producer interactions. A cell receptor can be specific for a cell of a particular type. At each time step a random index in the tissue compartment's cell store is chosen for each cell receptor. If a cell of the same type as the cell receptor exists at that index then that cell's index is copied to the receptor. Only when a cell is bound can certain other receptors, such as VR receptors, be activated.

VR receptors allow antigen presented on antigen producers to be matched. A VR receptor is the lock part of a lock-and-key type receptor mechanism. The lock is opened, that is the receptor activated, by certain antigen, the keys, which are presented on antigen producers of other cells. The exact structure of the locks and keys and the matching criteria chosen to establish which keys fit which locks is problem dependent. `libtissue` provides for this by allowing the user to specify the lock and key structure and matching in user-defined callback functions. VR receptors enable `libtissue` cells to perform antigen matching.

Antigen producers take antigen and make it available for inspection by other cells through VR receptors. Antigen producers work much like antigen receptors except that they transfer a randomly chosen antigen if available from a cell's internal store to the antigen producer itself. The antigen is removed from the cell's store and replaces any antigen which may already be on the antigen producer. This transfer and overwriting, when combined with antigen receptors, allows antigen to be passed through the system from tissue to cell to an antigen producer on a cell and so to eventual destruction. The parameter settings for the number of antigen receptors and producers a cell has, along with the size of the cell's antigen store, affect how quickly this process takes place. One further parameter is available which has proved useful in controlling this process. Antigen producers have an action time which determines the number of cell cycles an antigen is displayed for on the producer. While an antigen is being displayed, it cannot be overwritten by other antigen. Together with antigen receptors, antigen producers give a cell the ability to process antigen.

Cytokine producers allow signals stored in the tissue compartment to be set. At each time step the value on the cytokine producer affects the value to the corresponding cytokine in the tissue compartment. Since the values of cytokines can also be read by other cells, cells equipped with both cytokine receptors and producers are capable of signal processing and can form complex signalling networks. Response producers allow cells to send messages to `libtissue` response clients and so actively affect the systems they are monitoring. The semantics of the message and its actual effects are determined by user-supplied callbacks. In this paper, only a simple response producer which logs a message is considered. Active responses in combination with `libtissue` response clients are also possible. If the action of response producers is linked to cytokine and VR receptors in the cell cycle callback then cells can be made to respond to antigen in a selective way.

IV. AN EXAMPLE PROBLEM

The architecture described in the previous section allows AIS algorithms to be implemented and experimentally evaluated fairly easily, and an example algorithm will shortly be given. First, this section addresses how `libtissue` can be used to test algorithms on realistic data derived from real-world problems. A brief review of a real-world intrusion detection problem is now presented, followed by a short description of the datasets gathered for this problem.

Fundamentally, anomaly detection in intrusion detection rests on the idea of a normal profile of behaviour, deviations from which are considered as attacks [7]. It is attractive in that it allows novel attacks to be detected so long as one can determine to a sufficient degree of accuracy what is normal. Errors occur when instances of normal behaviour are seen as attacks, the false positive rate, or when attacks are seen as normal behaviour, the false negative rate. Reducing the false positive rate of anomaly detection systems is currently a key area of research in intrusion detection. Process anomaly detection is a specific example of one such anomaly detection problem. Several process anomaly detection systems have been built on the idea of using syscalls to monitor the behaviour of processes. Research such as [7] and [8] has shown that this avenue is promising, especially when combined with other sources of data such as context signals. Systems such as `systrace` [4] have also been implemented which allow process behaviour to be controlled through syscall policies.

In order to gather data for the process anomaly detection problem a small experimental network with three hosts was set up. Pairs of `strace` and `process_monitor` logs were collected on the instrumented target machine while `rpc.statd` was utilised in a number of different scenarios. These logs were then parsed to form a single `tcreplay` logfile for each of the scenarios. An antigen entry in the tcreplay log was created for every syscall recorded in the strace log. A signal entry was created for each recording of CPU usage in the process_monitor log. While the strace log actually contains much more information, the use of just the syscall number is more than sufficient for testing the example algorithm described in the next section. It would be expected that a more complex algorithm would require additional complexity in both the antigen and range of signals it is provided with, such as the addition information about syscall arguments, sequences of syscalls, or instruction pointer address. A larger number of datasets would also be necessary to statistically validate an algorithm. The monitored scenarios are divided into three groups based on whether the type of interaction with the rpc.statd server is a successful attack, a failed attack, or normal usage.

## V. AN EXAMPLE ALGORITHM

This section describes an example AIS algorithm called twocell implemented using libtissue. The algorithm is primarily intended to evaluate the initial libtissue implementation, and also as an explanatory aid to help the reader understand how the fairly complex system described in Section III is used to actually implement an algorithm. For this reason the functions and interactions of the cells in the example are kept fairly simple. This simplicity will of course limit the algorithm's overall performance on the problem when compared to existing solutions. On the other hand it allows for the behaviour of the algorithm to be traced at an in-depth level, the results of which are presented in Section VI below. This also makes the algorithm a useful tool for testing and evaluation of the libtissue implementation itself. Other papers such as [9] focus on more complex algorithms developed with libtissue.

### A. the twocell algorithm

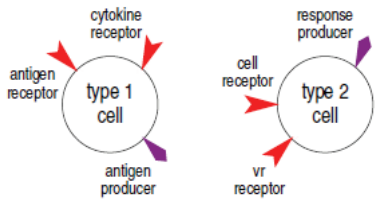

Fig. 2. The two different cell types implemented in twocell.

The cells in twocell, shown in Figure 2, are of two types, labelled Type 1 and Type 2, and each type has different receptor and producer repertoires, as well as different cell cycle callbacks. Type 1 cells are designed to emulate two key characteristics of biological APCs: antigen and signal processing. In order to process antigen, each Type 1 cell is equipped with a number of antigen receptors and producers. A cytokine receptor allows Type 1 cells to respond to the value of a signal in the tissue compartment. Type 2 cells emulate three of the characteristics of biological T cells:

cellular binding, antigen matching, and antigen response. Each Type 2 cell has a number of cell receptors specific for Type 1 cells, VR receptors to match antigen, and a response producer which is triggered when antigen is matched. Type 2 cells also maintain one internal cytokine, an integer which is incremented every time a match between an antigen producer and VR receptor occurs. If the value of this cytokine is still zero, that is no match has occurred, after a certain number of cycles, set by the cell_lifespan parameter, then the values of all of the VR receptor locks on the cell are randomised. Settings for the various parameters are given in Table I.

TABLE I
THE libtissue PARAMETER SETTINGS USED FOR twocell.

| | |
|---|---|
| max antigen | 1000 |
| max cytokines | 0 |
| max cells | 100 |
| cell update rate ($\mu$secs) | 100000 |
| antigen multiplier | 10 |
| num cells 1 | 50 |
| num antigen 1 | 100 |
| num antigen receptors 1 | 10 |
| num antigen producers 1 | 10 |
| antigen producer action time | 10 |
| num cells 2 | 50 |
| cell lifespan 2 | 100 |
| num cell receptors 2 | 2 |
| num vr receptors 2 | 20 |
| num response producers 2 | 1 |
| probe rate ($\mu$secs) | 1000000 |

A tissue compartment is populated with a number of Type 1 and 2 cells. Antigen and signals in the compartment are set by libtissue clients based on the syscalls a process is making and its CPU usage. Type 1 and 2 cells have different cell cycle callbacks. Type 1 cells ingest antigen through their antigen receptors and present it on antigen producers. The period for which the antigen is presented is determined by a signal read by a cytokine receptor on these cells, and so can be made dependent upon CPU usage. Type 2 cells attempt to bind with Type 1 cells via their cell receptors. If bound, VR receptors on these cells interact with antigen producers on the bound Type 1 cell. If an exact match between a VR receptor lock and antigen producer key occurs, the response producer on Type 2 cells produces a response, in this case a log entry containing the value of the matched receptor.

## VI. RESULTS

One of the goals of libtissue is to allow algorithms to be experimentally evaluated and tested. The aim of this section is to highlight through a handful of simple experiments the methodology employed when attempting to understand the dynamics of algorithms implemented with libtissue and when testing them on a real-world problem. twocell is used for this purpose and its behaviour is examined when applied to six datasets. The first experiment looks at a number of `twocell` runs, while the second takes one run and examines it more closely. The third evaluates the performance of a syscall policy generated by `twocell`. During these experiments, in order to more clearly understand the dynamics of `twocell`, the cytokine receptor on Type 1 cells is disabled, thus making `twocell` unresponsive to the CPU usage external signal. The final experiment returns to the question of signals and compares the effect the addition of the signal has on the dynamics of `twocell`. The parameters given in Table I were used for all experiments, which were carried out on a 2GHz AMD64 Turion laptop running Linux. Runs used on average around 1%, and never more than 3%, of the available CPU resources.

TABLE II

THE NAIVE SYSCALL POLICY AND THE AVERAGE `twocell` POLICY GENERATED FROM THE normal1 AND normal2 DATA SETS.

| syscall | freq | mean | sd | cv |
|---|---|---|---|---|
| chdir(12) | 2 | 0.07 | 0.26 | 371 |
| execve(11) | 2 | 0.07 | 0.26 | 371 |
| personality(136) | 2 | 0.07 | 0.34 | 485 |
| setsid(66) | 2 | 0.07 | 0.34 | 485 |
| fork(2) | 2 | 0.10 | 0.37 | 370 |
| write(4) | 2 | 0.10 | 0.37 | 370 |
| send(309) | 2 | 0.15 | 0.56 | 373 |
| time(13) | 2 | 0.15 | 0.40 | 266 |
| fstat64(197) | 2 | 0.17 | 0.52 | 305 |
| lseek(19) | 2 | 0.17 | 0.42 | 247 |
| fsync(118) | 2 | 0.25 | 0.80 | 365 |
| getrlimit(191) | 2 | 0.28 | 0.67 | 320 |
| listen(304) | 2 | 0.28 | 0.63 | 239 |
| select(142) | 3 | 0.57 | 1.48 | 225 |
| gettimeofday(78) | 4 | 0.50 | 0.85 | 276 |
| getsockname(306) | 4 | 0.53 | 1.47 | 170 |
| exit(1) | 4 | 0.55 | 1.38 | 277 |
| uname(122) | 4 | 0.75 | 1.91 | 250 |
| stat(106) | 4 | 0.80 | 2.58 | 259 |
| connect(303) | 5 | 1.60 | 2.48 | 254 |
| getdents(141) | 8 | 0.20 | 0.73 | 322 |
| mprotect(125) | 8 | 0.47 | 1.30 | 185 |
| poll(168) | 8 | 0.90 | 1.67 | 224 |
| sendto(311) | 9 | 0.95 | 2.13 | 225 |
| recvfrom(312) | 9 | 2.45 | 3.68 | 233 |
| rt_sigaction(174) | 10 | 0.97 | 2.19 | 155 |
| getpid(20) | 10 | 1.60 | 2.28 | 142 |
| fcntl(55) | 12 | 1.18 | 2.76 | 268 |
| bind(302) | 12 | 1.68 | 4.51 | 200 |
| munmap(91) | 15 | 1.88 | 3.77 | 225 |
| brk(45) | 16 | 2.25 | 3.78 | 168 |
| fstat(108) | 23 | 2.33 | 4.45 | 229 |
| ioctl(54) | 24 | 2.73 | 4.67 | 190 |
| socket(301) | 25 | 3.10 | 4.97 | 150 |
| old_mmap(90) | 27 | 1.90 | 4.29 | 171 |
| read(3) | 27 | 2.25 | 5.17 | 160 |
| open(5) | 30 | 5.95 | 7.75 | 130 |
| close(6) | 557 | 19.43 | 27.03 | 139 |

In experiments it is important to have a baseline with which to compare algorithmic performance. In terms of syscall policies such a baseline can be generated and is here termed a *naive policy*. A naive syscall policy is generated for a process, such as `rpc.statd`, by recording the syscalls

TABLE III

THE SYSCALL POLICY GENERATED BY `twocell` FOR THE normal2 DATA SET AND THE FREQUENCY OF RESPONSE FOR EACH SYSCALL.

| syscall | frequency |
|---|---|
| gettimeofday(78) | 1 |
| listen(304) | 1 |
| send(309) | 1 |
| select(142) | 2 |
| poll(168) | 3 |
| recvfrom(312) | 8 |
| fcntl(55) | 9 |
| fstat(108) | 9 |
| open(5) | 22 |
| close(6) | 34 |

it makes under normal usage, as in the normal1 and normal2 datasets. A permit policy statement is then created for all syscalls seen. This baseline is not too unrealistic when compared to how current systems such as `systrace` automatically generate a policy. The first column of Table II shows the permitted syscalls (syscall number given in brackets) in such a naive policy generated from the normal1 and normal2 datasets. The frequency with which each syscall was observed at combined over the two datasets is given in the second column, as this will be useful for further analysis.

Similarly to the naive policy, one way in which `twocell` can be used is to generate a syscall policy by running it with normal usage data during a training phase. During the run, responses made by Type 2 cells are recorded. At the end of each run, a syscall policy is created by allowing only those syscalls responded to, and denying all others. Since interactions in `libtissue` are stochastic, looking at the average results over a number of runs helps to understand the behaviour of implemented algorithms. A script was written to start the `twocell` server and then after 10 seconds start the `tcreplay` client and replay a dataset in realtime. `twocell` was allowed to continue running for a further minute after replay had finished. This process was repeated 20 times for both the normal1 and normal2 datasets, yielding 40 individual syscall policies. A single average `twocell` policy was then generated by allowing all syscalls which were permitted in any of the 40 individual policies. It was found that all of the 38 syscalls that were permitted in the naive policy were also permitted in the average policy. The mean frequency with which the syscall appeared in a policy is given in the third column of Table II. As expected, there appears to be a correlation between the frequency that a syscall occurs and the likelihood of it being in a policy generated by `twocell`. Standard deviations, given in the fourth column of Table II, appear to at first show an increasing amount of noise for high-frequency syscalls. However, examination of the coefficient of variation for each syscall, given in the last column of Table II, shows that there is in fact more variation in the frequencies of response to the lower frequency syscalls.

The last experiment showed that the `twocell` algorithm has the property of responding in a selective way to input data based on the frequency at which an input data item occurs. In order to examine more closely how twocell responds, a single run of the twocell algorithm was observed. Following the same general procedure as the previous experiment, twocell was run once with the normal2 dataset. The resulting policy is shown in Table III, along with the frequencies with which the permitted syscalls were responded to. During the run, the time at which a Type 2 cell produced a response to a particular syscall was also recorded, and the rate at which these responses occur is plotted in Figure 3. The rate of incoming syscalls is also plotted for comparison. This figure clearly shows a correlation between the rate of incoming syscalls and the rate of responses produced by Type 2 cells. Cells initially do not produce any response until syscalls occur, and then produce a burst of responses for a relatively short period before settling down to an unresponsive state once again. This is to be expected, as antigen enter and are passed through twocell until their eventual destruction after being presented on Type 1 cell antigen producers.

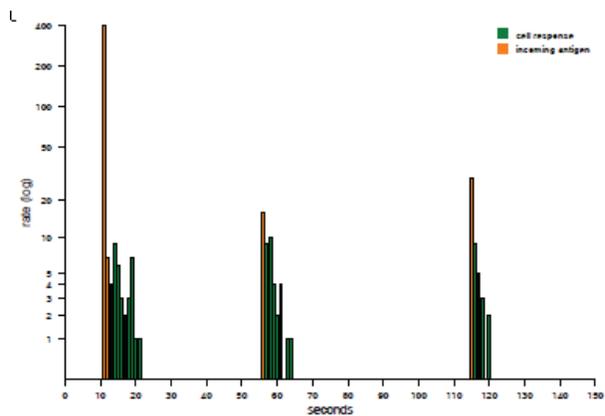

Fig. 3. The rate of incoming antigen and corresponding cell response rates for the normal2 dataset.

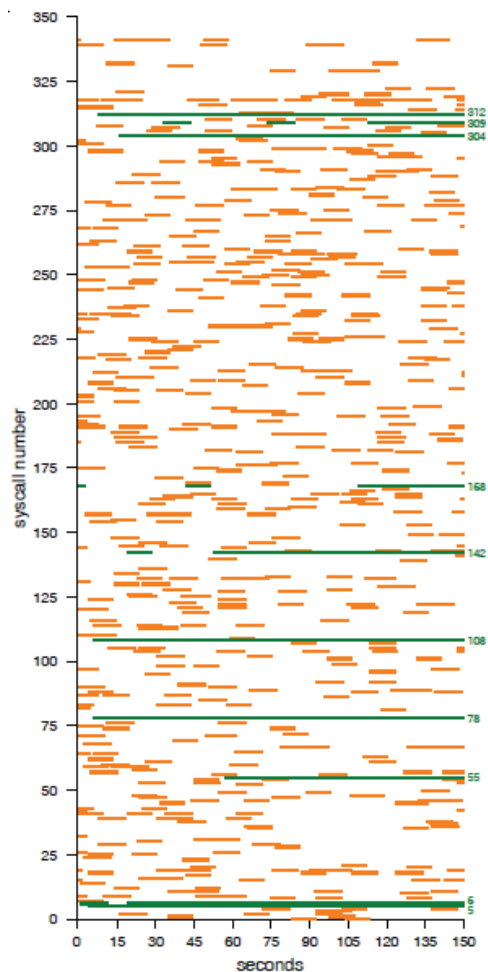

Fig. 4. The VR receptor repertoire expressed by Type 2 cells for the normal2 dataset. Highlighted syscalls are the ones responded to.

For the same run, the individual receptors expressed by Type 2 cells can also be examined. Figure 4 shows the repertoire of VR receptors expressed by all 50 Type 2 cells during the run. A libtissue probe periodically recorded the syscall values expressed by the VR receptors on all of the Type 2 cells. A point is plotted in Figure 4 if the syscall was being expressed during that period. Points for the 10 syscalls which twocell responded to (see Table III) are highlighted. As expected, due to the limited lifespan of unmatched Type 2 cells, set by the cell_lifespan parameter, and after which the cell's VR receptor is randomised, many bursts of around 10 seconds of expression of VR receptors specific for a given syscall are seen. Once a VR receptor matches, and a response and permit policy is therefore produced for that syscall, the cell stops randomising its receptors. This can be observed from the continuous horizontal lines in Figure 4 for the 10 highlighted syscalls.

An example is now given of how the classification accuracy and error of a libtissue algorithm can be evaluated. In terms of syscall policies, a particular policy can be considered successful in relation to the number of normal syscalls it permits versus the number of attack syscalls it denies. The naive policy and average twocell policy generated from datasets normal1 and normal2 in the first experiment above were evaluated in such a way. The number of syscalls both policies permitted and denied when applied to the four datasets in the attack and failed groups was recorded. For each dataset, Table IV shows the percentages of attack and normal syscalls in the dataset, together with the percentage of syscalls permitted by the naive and twocell policies. The results show that the tendency of the naive policy was to permit the vast majority of syscalls, whether attack related or not. The twocell generated policy behaved much more selectively, denying a slightly larger proportion of syscalls in the success1 and success2 datasets than it permitted. For the failure1 and failure2 datasets the converse was true.

The previous experiments have all used the twocell al-gorithm with the cytokine receptors of Type 1 cells disabled. This was necessary in order to gain an initial understanding of the dynamics of twocell. This final experiment now examines how the addition of a context signal changes the dynamics of the algorithm. When enabled, the cytokine receptor on a Type 1 cell controls the action_time parameter of antigen producers on these cells as follows. The action_time parameter is initialised to a value of 100. If there is no change in the signal, CPU usage in this case, then the action time stays the same. If CPU usage decreases, the action time is reduced by 50%, and if it increases, the action time is reset 100. twocell with its cytokine receptor enabled was run 20 times on the success2 dataset and the responses it produced recorded. For a fair comparison, the mean action time observed on antigen producers over all of the runs, 28.57 in this case, was calculated and the twocell algorithm *without* signals was run 20 times on the same dataset with the action time of its antigen producers set to 29. Figure 5 shows bspline curves fitted to the mean response rates of twocell with and without a signal over the 20 runs. The results show that the response time of twocell with a signal is much more tightly controlled, with responses starting and dropping off more rapidly and lasting for a shorter duration in total. This is to be expected in light of the incoming data, and from the action of the cytokine receptor, which causes a sudden rise and quick decreases in the action time of the antigen producers on Type 1 cells based on the rate of change of the external signal.

TABLE IV
COMPARISON OF THE PERFORMANCE OF A NAIVE POLICY AND A
twocell POLICY GENERATED FROM THE normal2 DATA SET.

| dataset | success1 | success2 | failure1 | failure2 |
|---|---|---|---|---|
| normal syscalls | 23% | 23% | 81% | 87% |
| attack syscalls | 76% | 76% | 18% | 12% |
| naive permit | 90% | 90% | 99% | 99% |
| naive deny | 9% | 9% | 0% | 0% |
| twocell permit | 47% | 47% | 69% | 68% |
| twocell deny | 52% | 52% | 30% | 31% |

## VII. CONCLUSIONS

The aim of this paper has been to describe the architecture of the libtissue implementation and how it is used to implement and evaluate algorithms on real-world problems. After briefly laying down the biological and conceptual background, the libtissue implementation was described in detail. In order to help understand how libtissue is actually used, its application to a real-world intrusion detection problem was presented. An example algorithm implemented with libtissue was then introduced, and aspects of its dynamics evaluated and discussed. The paper now concludes with a brief summary and discussion of future work.

While simplified, the examples presented above validate the libtissue implementation in several ways. They show that it meets the goals it set out to achieve in terms of im-plementation, evaluation and application of AIS algorithms. More generally, they show the feasibility of using AISs implemented as multiagent systems to address real-world problems. Additionally, it is the authors' experience that simple algorithms such as twocell are a necessary step in developing more complex algorithms. Such algorithms are being developed by the authors and other researchers using libtissue and future papers will report the results of this research. The sourcecode of libtissue is distributed under a GPL licence and available, along with the datasets, clients and example algorithm used in this paper, from the first author's website.

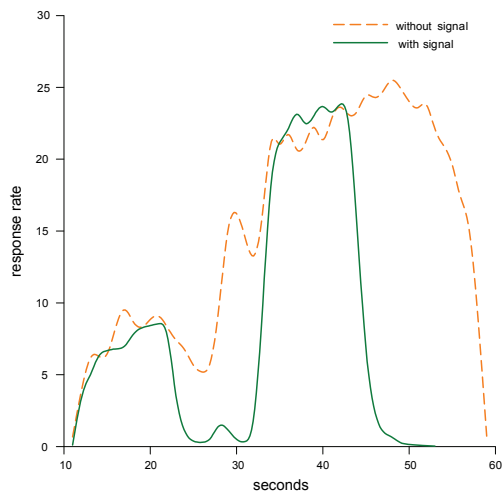

Fig. 5. The mean response rates of the twocell algorithm with and without a signal for 20 runs on the success2 dataset.

ACKNOWLEDGMENTS

This research is supported by the EPSRC (GR/S47809/01).